\documentclass[journal]{IEEEtran}
\usepackage[pdftex]{graphicx}
\ifCLASSINFOpdf
\else
\fi
\usepackage{amsmath}
\usepackage{caption}
\usepackage{subcaption}
\usepackage{cite}
\usepackage{multirow}
\usepackage{verbatim}
\usepackage{float}
\usepackage{dblfloatfix}
\usepackage{booktabs}

\begin{document}

%
\title{Dual-Modality Haptic Feedback Improves Dexterous Task Execution with Virtual EMG-Controlled Gripper}
%
%
%

\author{Kezi Li,
        Jeremy D. Brown,~\IEEEmembership{Member,~IEEE}
\thanks{K. Li was with the Laboratory for Computational Sensing and Robotics, Johns Hopkins University, Baltimore, MD, USA (e-mail: \mbox{kli68@jhu.edu})}  
\thanks{J. Brown is with the Department
of Mechanical Engineering, Johns Hopkins University, Baltimore,
MD, USA (e-mail: \mbox{jdelainebrown@jhu.edu})}}

\maketitle



\begin{abstract}

Upper-extremity amputees who use myoelectric prostheses currently lack the haptic sensory information needed to perform dexterous activities of daily living. While considerable research has focused on restoring this haptic information, these approaches often rely on single-modality feedback schemes which are necessary but insufficient for the feedforward and feedback control strategies employed by the central nervous system. Multi-modality feedback approaches have been gaining attention in several application domains, however, the utility for myoelectric prosthesis use remains unclear. In this study, we investigated the utility of dual-modality haptic feedback in a virtual EMG-controlled grasp-and-hold task with a brittle object and variable load force. We recruited N=20 non-amputee participants to perform the task in four conditions: no feedback, vibration feedback of incipient slip, squeezing feedback of grip force, and dual (vibration + squeezing) feedback of incipient slip and grip force. Results suggest that receiving any feedback is better than receiving none, however, dual-modality feedback is far superior to either single-modality feedback approach in terms of preventing the object from breaking or dropping, even after it started slipping. Control with dual-modality feedback was also seen as more intuitive than with either of the single-modality feedback approaches.

\end{abstract}


%
\IEEEpeerreviewmaketitle

\section{Introduction}

Our ability to dexterously perform activities of daily living is predicated on the availability of haptic sensations streaming from the myriad of cutaneous and kinesthetic sensory receptors in our peripheral limbs \cite{Johansson1987SignalsGrip,Johansson2009CodingTasks}. These haptic cues are used by the central nervous system to track task progression through feedback control and update internal models needed for predictive feedforward control \cite{Johansson1984RolesObjects,Augurelle2003ImportanceObjects}. Together, this feedforward and feedback sensorimotor control loop allows for the execution of object manipulation tasks that require accurate prediction of object properties, as well as robust compensatory strategies for task errors or uncertainties \cite{Johansson1988ProgrammedGrip}.


When a limb of the upper-extremity is amputated and replaced with a myoelectric prosthesis, this haptic sensory feedback channel remains incomplete, forcing users to rely heavily on visual and auditory cues to guide prosthesis usage \cite{Cordella2016LiteratureUsers}. Vision has been shown to carry a high cognitive burden \cite{Biddiss2007Upper-limbAbandonment,Thomas2021NeurophysiologicalProstheses} and limits the amount of visual information that can be used for task planning \cite{Sobuh2014VisuomotorProsthesis}. Furthermore, visual information has been shown to be inferior to haptic information in dexterous task performance with a body-powered prosthesis, which features inherent force feedback \cite{Brown2017AnProstheses}. Several research studies have attempted to address this shortcoming through the development of novel haptic feedback approaches that provide a range of haptic cues including grip force, grip aperture, and incipient slip \cite{Witteveen2015VibrotactileUsers,Damian2012SlipControl,Thomas2021Sensorimotor-inspiredVision,Wheeler2010InvestigationSystems}. Unfortunately, the majority of these efforts have focused on a single form of haptic feedback, which still limits the overall amount of haptic information the user is receiving. 

\begin{figure}[t]
    \centering
    \includegraphics[width=\columnwidth]{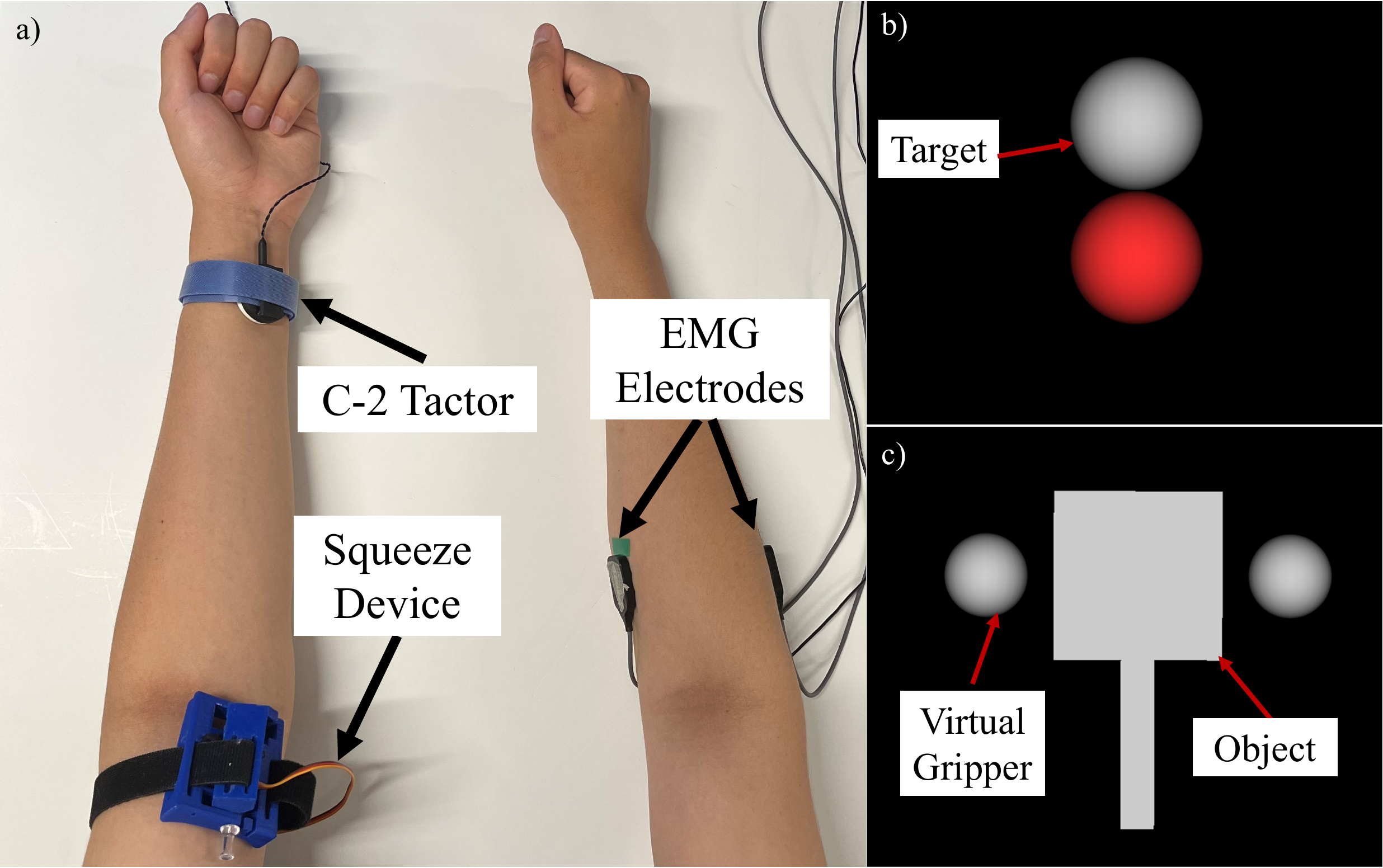}
    \caption{a) The C-2 tactor and squeeze device were placed on the participant's non-dominant arm. The EMG electrodes were placed on the participant's dominant arm. b) EMG practice task virtual environment c) Grasp-and-hold task virtual environment }
    \label{fig:devices_VE}
    \vspace{-3.5mm}
\end{figure}

In an effort to overcome the shortcomings of single-modality haptic approaches, recent research has demonstrated the potential utility of multi-modality feedback for a range of haptic interaction and telerobotic applications \cite{Lim2015RoleSurgery,Sullivan2020Multi-SensoryCues,Abiri2019ArtificialFeedback,Abiri2019Multi-ModalSurgery,Santos-Carreras2011MultimodalRobotics,Machaca2020TowardsDevices,Machaca2022TowardsAssessments}. For prosthetic applications in particular, Walker et al. discovered that force feedback of object contact combined with vibrotactile feedback of object slip resulted in less object slips and more object recoveries than single-modality feedback or no feedback \cite{Walker2015TactileManipulation}. Likewise, Xu et al. demonstrated that electrotactile feedback of grasping pressure and incipient slip results in performance that is just as good as visual feedback \cite{Xu2016EffectsStimulation}. 

Despite these benefits, other research has found that multi-modality feedback is not as helpful. For example, Kim et al. showed that dual-modality haptic feedback (pressure and shear) degrades grip force control more than single-modality feedback for myoelectric prostheses \cite{Kim2012HapticAmputees}. Likewise, Jimenez et al. showed that multi-modality (force, vibration, and thermal feedback) could be distracting for participants controlling a prosthesis \cite{Jimenez2014EvaluationLimbs}. It should be noted, however, that both of these studies only involved a single amputee participant. Likewise, they did not explicitly investigate the feedback's utility in helping the user develop closed-loop control strategies that are robust to task uncertainties and task errors. 
Thus, it is still unclear if, and to what extent, multi-modality haptic feedback provides added utility over single-modality feedback for myoelectric prosthesis usage. In particular, it is worth investigating to what degree multi-modality approaches support the feedforward and feedback sensorimotor control approaches utilized by the central nervous system in dexterous task execution. 

In this manuscript, we present the findings of a user-study designed to investigate whether providing continuous dual-modality haptic feedback of grip force and incipient slip can improve myoelectric prosthesis users' ability to perform a unique dexterous manipulation task. Building on the widely accepted grasp-and-lift task literature \cite{Flanagan2006ControlTasks,Johansson1988CoordinatedGrip,Johansson1988ProgrammedGrip,Brown2015AnFeedback}, we ask participants to perform a virtual grasp-and-hold task with a brittle object that experiences an increasing but invisible load force. Using an EMG-controlled virtual gripper, participants attempt to complete the task without visual feedback in four conditions: no feedback, vibration feedback of incipient slip, squeezing feedback of grip force, and dual (vibration + squeezing) feedback of incipient slip and grip force. While the use of a virtual task environment does not exactly replicate real-world prosthesis usage, it removes the confounding factors of physical fatigue and poor EMG control associated with real prosthesis usage. We \underline{hypothesize} that the dual-modality feedback condition will result in the best task performance given that it provides participants with the necessary and sufficient haptic information needed to successfully keep the virtual object from breaking or dropping. In what follows, we describe our experimental method, followed by a presentation of our experimental findings, and a discussion of our results in the context of existing multi-modality literature. 

%
%
%
%
\section{Methods}
We recruited n=20 non-amputee participants (12 male and 8 female, age 22$\pm$3 years, 17 right-handed) to perform a virtual grasp and hold task using surface electromyography (EMG) control. All participants were consented according to a protocol approved by the Johns Hopkins University IRB (HIRB00005942). The duration of the experiment was approximately 90 min, and participants were compensated at a rate of \$10/hour.

\subsection{Experimental Tasks}
There were two experimental tasks, an EMG practice task and a grasp and hold task. Both tasks were performed in a virtual environment as described below: 

\subsubsection{EMG Practice Task} \label{section:sEMG Calibration and Processin}
The EMG practice task was designed to help participants become comfortable with EMG control and provided a quantifiable baseline measure of participants' EMG control capabilities. In the task (see Fig. \ref{fig:devices_VE}b) participants are asked to control the horizontal velocity of a virtual ball (red) using EMG to track the horizontal position of a target ball (gray), which moves in a sinusoidal manner. The target ball's position was governed by the following two equations, which created two different levels of difficulty:
\begin{equation}
  X_{target}(t) =
    \begin{cases}
      2\cdot sin(t) & \text{easy}\\
      
      \frac{3}{4}\cdot sin(\frac{t}{3}) + \frac{3}{4}\cdot sin(\frac{4t}{3}) & \text{hard}
    \end{cases}       
\end{equation}
\noindent where $t$ is the elapsed time since the start of the trial and $X_{target}$ is the horizontal position of the target sphere from the center. 

\subsubsection{Grasp-and-Hold Task}  \label{section:grasp-and-hold task}
The grasp-and-hold task was designed to test the utility of different forms of haptic feedback on performance in a dexterous task. In the task, the participants' objective was to grasp and hold the virtual object with the virtual gripper, as shown in Fig. \ref{fig:devices_VE}c. Participants controlled the velocity of the virtual grippers using EMG control. During the task, the object was pulled toward the bottom of the screen by a logarithmically increasing pulling force (not visible), as described by
\begin{equation}\label{eqn:pulling}
  F_{pull}(t) =K_{f}\cdot \ln(t+1)
\end{equation}
\noindent Where $K_f$ = [1,2,3,3.5,4,5] that controls the intensity of the pulling force, and t is the elapsed time since the start of the trial. In addition to the pulling force, the virtual object was ``brittle," with a breaking force that was computed as a constant offset above the minimum required grip force needed to hold the object:
\begin{equation}\label{eqn:breaking_threshold}
  F_{break} = F_{pull}(t) + 4
\end{equation}
\noindent where $F_{pull}(t)$ is from Eq. \ref{eqn:pulling}. The constant offset was determined heuristically through pilot testing and produces a grip force safe region wherein the object won't slip or break.

\paragraph{Task procedure} There was a five-second preparation period before each trial started. During the preparation period the virtual object was not pulled and participants were asked to move the virtual grippers until they touched the virtual object. After the preparation period, the trial started with the virtual object disappearing and the pulling force increasing. The trial ended if the participant successfully held the object for five seconds. In addition, the trial ended prematurely if the participant broke or dropped the object. If the participants held the object successfully for five seconds, the object reappeared in the original position. If the participant successfully recovered the object from slipping, the object reappeared in the corresponding position where the object was recovered. If participants broke the object, it reappeared as red in the original position or the corresponding position where it slipped before breaking occurred. If the object was dropped, it reappeared as red and fell to the bottom of the screen.  

\subsection{Experimental Setup}
The experimental setup consisted of the virtual task environment (described above), haptic feedback display, and a Delsys Bagnoli 8-channel sEMG system to collect surface EMG signals from the user's wrist flexor and extensor muscles. The magnitude of the muscle activity was mapped proportionally to the horizontal velocity of the red ball in the practice task (see section \ref{section:sEMG Calibration and Processin}), and to the opening and closing velocity of the virtual grippers in the grasp and hold task (see section \ref{section:grasp-and-hold task}). Data acquisition and control were implemented with an NI myRIO at 1\,kHz with MATLAB/Simulink (2021a) and QUARC real-time software (2021 v4.1.3406). The virtual environment (VE) was displayed at 100 Hz, based on down-sampled data. The entire system ran on a Dell Precision Tower 3620 desktop with Windows 10.

\subsubsection{sEMG Calibration and Processing}\label{sEMG Calibration and Processin}
The EMG calibration method followed the same procedure used in previous work by Thomas et al \cite{Thomas2019ComparisonProsthesis}. First, participants were asked to place their arm on a flat armrest with their wrist in a neutral position to collect their baseline EMG signals. Then, participants were asked to produce a series of maximum voluntary contractions (MVCs) of their flexor muscles for eight seconds. This MVC procedure was then repeated for the extensor muscles. The threshold percentage for flexion depended on the ratio of averaged flexor peaks during extension to the averaged flexor peaks during flexion, as shown in:

\begin{equation}\label{eqn:p_flexion}
  P_{flexion} =
  \left(\frac{F_{extension}}{F_{flexion}} + 0.1\right) \cdot 100 \\
\end{equation}

\noindent where $F_{extension}$ is the averaged flexor peaks during extension and $F_{flexion}$ is the averaged flexor peaks during flexion. Likewise, the threshold percentage for extension depended on the ratio of averaged extensor peaks during flexion to the averaged extensor peaks during extension, as shown in:  

\begin{equation}\label{eqn:p_extension}
  P_{extension} =
  \left(\frac{E_{flexion}}{E_{extension}} + 0.1\right) \cdot 100
\end{equation}

\noindent where $E_{flexion}$ is the averaged extensor peaks during flexion, and $E_{extension}$ is the averaged extensor peaks during extension.

The upper threshold for flexion $F_{up\_thr}$ and extension $E_{up\_thr}$, and the lower threshold for flexion $F_{low\_thr}$ and extension $E_{low\_thr}$ were calculated as: 

\begin{equation}\label{eqn:up_flex_thre}
F_{up\_thr} =
F_{flexion}\cdot P
\end{equation}

\begin{equation}\label{eqn:up_ext_thre}
E_{up\_thr} =
E_{extension} \cdot P
\end{equation}

\begin{equation}\label{eqn:low_flex_thre}
F_{low\_thr} =
\frac{F_{extension}}{F_{flexion} \cdot P}
\end{equation}

\begin{equation}\label{eqn:low_ext_thre}
E_{low\_thr} =
\frac{E_{flexion}}{E_{extension} \cdot P}
\end{equation}

\noindent where $F_{extension}$, $F_{flexion}$, $E_{flexion}$, and $E_{extension}$ are the same as above, and 

\begin{equation}
    P = max(P_{flexion},P_{extension})
\end{equation}

Using the respective upper and lower thresholds for both flexion and extension, the normalized flexor $S_{flex}(t)$ and extensor $S_{ext}(t)$ signals were calculated as:

\begin{equation} \label{eqn:S_flex}
\begin{split}
  S_{flex}(t) &=
    \begin{cases}
    0  & \text{ $ F_{user} < F_{low\_thr} $} \\
     \frac{F_{user}}{F_{up\_thr}} & \text{ $F_{low\_thr} < F_{user} < F_{up\_thr}$} \\
      1 & \text{$ F_{user} > F_{low\_up} $} \\
    \end{cases} \\
\end{split}
\end{equation}

\begin{equation} \label{eqn:S_ext}
\begin{split}
  S_{ext}(t) &=
    \begin{cases}
    0  & \text{ $ E_{user} < E_{low\_thr} $} \\
     \frac{E_{user}}{E_{up\_thr}} & \text{ $E_{low\_thr} < E_{user} < E_{up\_thr}$} \\
      1 & \text{$ E_{user} > E_{up\_thr} $} \\
    \end{cases} \\
\end{split}
\end{equation}

\noindent where $F_{user}$ and $E_{user}$ are participant's flexor and extensor EMG signals, respectively, and $F_{up\_thr}$, $E_{up\_thr}$,$F_{low\_thr}$ and $E_{low\_thr}$ are the same as above.

Finally, the saturated net EMG signal $S_{net}{t}$ was calculated using the normalized flexor and extensor signals as:

\begin{equation}\label{eqn:Normalized EMG}
  S_{net}(t) = S_{flex}(t) - S_{ext}(t)
\end{equation}

\noindent $S_{net}(t)$ was used to control the velocity of the user controlled sphere, $v_{ball}(t)$, in the practice task (see section \ref{section:sEMG Calibration and Processin}) according to the following control law: 

\begin{equation}\label{eqn:V_ball}
  v_{ball}(t) = sign(h) \cdot K_{emg} \cdot S_{net}(t)
\end{equation}

\noindent where $K_{emg}$ is a positive gain used to scale the EMG signals, $sign(h)$ is the handedness of the participant with 1 for left-handed participants and -1 for right-handed participants.  

Likewise, $S_{net}(t)$ was used to control the opening and closing velocity of the virtual grippers, $v_{grippers}(t)$, in the grasp and hold task (see section \ref{section:grasp-and-hold task}) according to the following control law:

\begin{equation}\label{eqn:V_grippers}
  v_{grippers}(t) = K_{emg} \cdot S_{net}(t)
\end{equation}

\noindent where $K_{emg}$ is the same as in Eq \ref{eqn:V_ball}.


\subsection{Experimental Protocol } 
\begin{figure*}[t!]
    \centering
    \includegraphics[width=\textwidth]{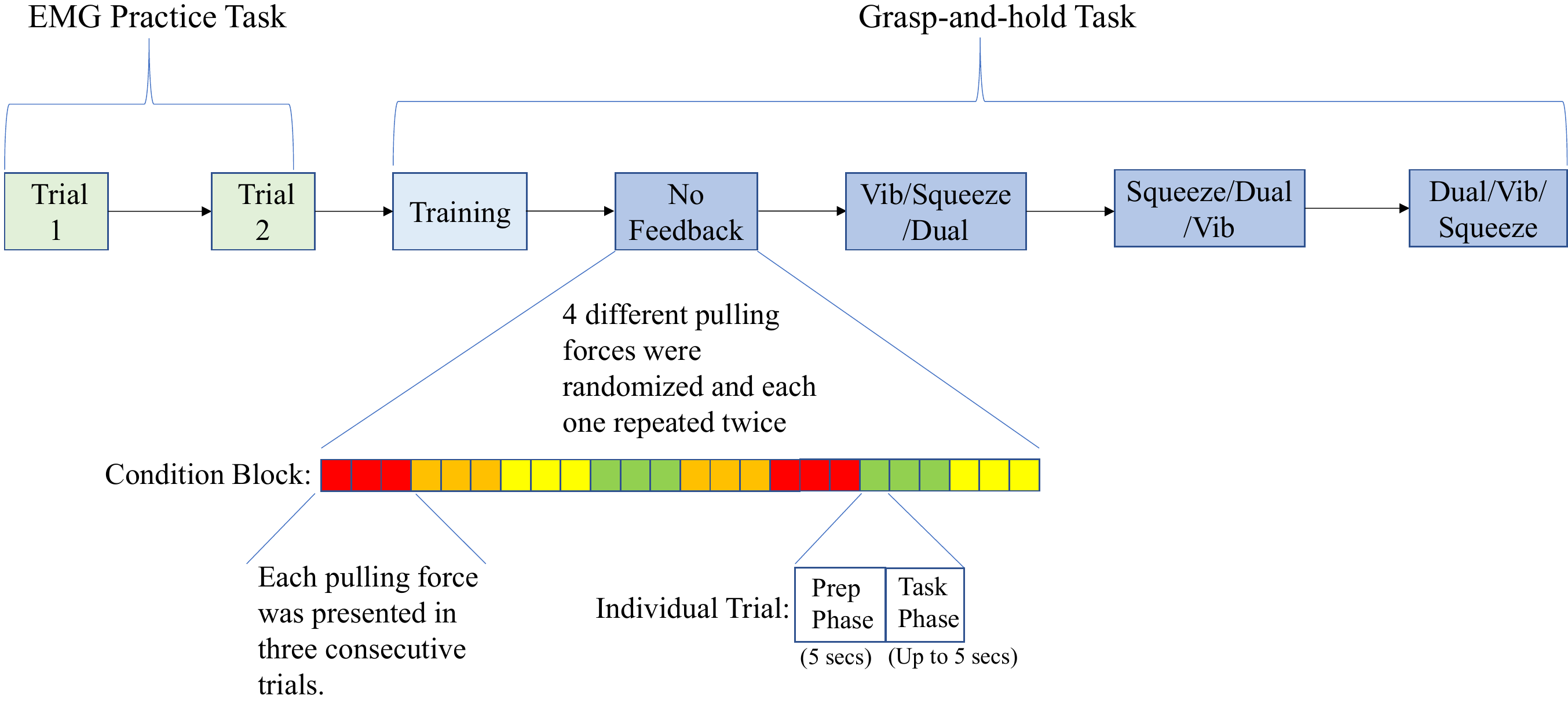}
    \caption{Experimental protocol design. All participants completed two practice task trials following by a training session for the grasp-and-hold task. All participants started with the no feedback condition, then the order of the remaining three feedback conditions was randomized and counterbalanced. For each condition, four different pulling forces were presented twice, with each pulling force presented in three consecutive trials, resulting in 24 trials per condition (96 trials for the entire session). }
    \label{fig:protocol}
    \vspace{-3.5mm}
\end{figure*}

\subsubsection{Haptic Feedback}  
\label{section:Haptic Feedback}
In the experimental task, haptic feedback in the form of vibration and squeeze was used to provide participants with knowledge of object slip and grip force, respectively.

\paragraph{Vibrotactile Feedback}
Vibrotactile feedback was provided by a C-2 tactor (Engineering Acoustics, Inc.), and it was placed in a custom 3-D printed housing. The C-2 tactor was controlled using a Syntacts \cite{Pezent2021Syntacts:Haptics} amplifier board controlled through Simulink. The vibration feedback frequency was fixed at 400 Hz, and the amplitude of the vibration, $a_amp(t)$, was modulated according to the slip velocity of the virtual object as shown in:
\begin{equation}
\begin{split}
  a_{amp}(t) &=
    \begin{cases}
      \beta(t) & \text{ $V_{slip}(t) >0 $}\\
      0 & \text{Otherwise}
    \end{cases} \\
    \beta(t) &= \frac{200\cdot \ln(2 \cdot |V_{slip}(t)| + 0.6)+102}{55}
    \label{eqn:vib}
\end{split}
\end{equation}

\noindent where $V_{slip}(t)$ is the slip velocity of the virtual object. The other constants were determined heuristically during pilot testing as described in \cite{Machaca2020TowardsDevices}. 

\paragraph{Squeezing Feedback} 
The squeezing display was provided using a custom device based on work by Stanley and Kuchenbecker \cite{Stanley2012EvaluationGuidance}. An MG90s servo was used to pull on the Velcro strap wrapped around the participant's arm to provide a squeezing sensation in proportion to the grip force that the participant applied to the virtual object. An Arduino micro-controller and Simulink were used to control the device. The angle of the servo (in degrees), $\theta(t)$, was given by
\begin{equation}
\theta(t) = K_{squeeze} \cdot F_{grip}(t) + \theta_{0}
\end{equation}

\noindent where $F_{grip}(t)$ is the grip force that participants applied to the virtual object, $K_{squeeze}$ is a constant coefficient that is used to map the maximum grip force to the maximum angle for which the feedback remained  comfortable, $\theta_{0}$ is the constant starting angle of the servo when $F_{grip}(t) = 0$. Both $K_{squeeze}$ and $\theta_{0}$ values were adjusted for each participant to ensure a full and comfortable range of sensations were felt. 

The C-2 tactor was placed on the participant's non-dominant wrist, and the squeeze device was placed around the participant's non-dominant bicep, as shown in Fig. \ref{fig:devices_VE}a. This decision came as the result of pilot studies, wherein participants mentioned that it would take more mental effort to feel the feedback if the haptic displays were placed on the same arm that participants used to do the EMG control. Also, participants mentioned that the vibration intensity overpowered the squeezing intensity if the C-2 tactor and the squeeze device were placed next to each other on the arm. 

\subsubsection{User study Procedure}
After consenting to participate in the study, participants were given an overview of the study procedures and asked to complete a demographic survey. Next, the experimenter placed the EMG electrodes on the participants' arm to do the EMG calibration, as detailed in Section \ref{sEMG Calibration and Processin}. Participants then performed two 30-second trials of the EMG practice task at each of the two difficulty levels as detailed in Section \ref{sEMG Calibration and Processin}. After the EMG practice tasks, the experimenter placed the C-2 tactor and squeeze device on the participants, as shown in Fig. \ref{fig:devices_VE}a. Participants were then asked to watch a video demonstrating the experimental procedure, including the preparation and task phases for each trial. The video also demonstrated the visual feedback for each of the possible outcomes at the end of the trial: a broken object, a dropped object, or a successful trial. After watching the demonstration video, participants performed six practice trials for each feedback condition to become better familiar with the task and haptic cues. In the  practice trials, participants were given two pulling forces, $K_{pull}$ = 1, 3.5 in equation \ref{eqn:pulling}. Participants received haptic feedback for each pulling force in the following order: no feedback, vibrotactile feedback only, squeezing feedback only, and both vibrotactile feedback and squeezing feedback. Participants were able to see the virtual object during the practice trials. After completing the practice trials, the main experiment began. All participants started the main experiment in the no feedback condition. The remaining three haptic feedback conditions were randomized and counterbalanced to minimize ordering affects. For each haptic feedback condition, four pulling forces were used, $K_{pull}$ = [2,3,4,5] in equation \ref{eqn:pulling}. Each of the four pulling forces was presented twice as a sequence of three consecutive trials (24 trials), in a randomized manner. In this way, participants had an opportunity to learn trial-by-trial the appropriate amount of grip force needed for each new set of object parameters (i.e., pulling force and breaking threshold) before the parameters changed. After completing each condition, participants completed a short survey based on the NASA-TLX assessment. An illustration of the overall study design is shown in Fig.\ \ref{fig:protocol}.

\subsection{Metrics}
The following metrics were used to analyze participants' performance in the practice task and the grasp-and-hold task. 
\subsubsection{EMG Practice Task Metrics}
Position root-mean-square error $P_{RMSE}$ and jerk root-mean-square error $J_{RMSE}$ were used to analyze the EMG practice task data. $P_{RMSE}$ measures the accuracy of each participant's EMG control by comparing the RMSE between the position of the target sphere and the participant's sphere as shown in  
 \begin{equation}\label{eqn:position_RMSE}
  P_{RMSE} =
  \sqrt{\sum_{t=0}^{30}({x}_{part}(t) - {x}_{target}(t))^2 }
\end{equation}
\noindent where ${x}_{part}(t)$ and ${x}_{target}(t)$ are the positions of the participant and target spheres at time $t$, respectively. 

$J_{RMSE}$ measures the smoothness of each participant's EMG control by comparing the RMSE between the jerk of the target sphere and the participant's sphere as shown in
\begin{equation}\label{eqn:jerk_RMSE}
  J_{RMSE} =
  \sqrt{\sum_{t=0}^{30}(\dddot{x}_{part}(t) - \dddot{x}_{target}(t))^2 }
\end{equation}
\noindent where $\dddot{x}_{part}(t)$ and $\dddot{x}_{target}(t)$ are the jerk of the participant and target sphere at time $t$, respectively. 

\begin{table}[b!]
\centering
    \small
    \caption{Fixed effect of EMG practice task metrics on grasp-and-hold task performance metrics for all trials}
    \label{table:EMG experience}
    \renewcommand{\arraystretch}{1.35}
    \begin{tabular}{l l c c c}
        \hline \hline
      metrics & EMG experience & $\beta$ & \textit{SE} & p \\ [0.5mm]
      \hline 
    \multirow{2}{*}{\#broken}&position RMSE &  0.977  & 0.816  & 0.231    \\
    &jerk RMSE   &  -0.293  & 0.399 & 0.462         \\
\cline{1-5}

        \multirow{2}{*}{\#dropped}&position RMSE &   0.346  & 1.531   & 0.821      \\
    &jerk RMSE   &  -0.121 &  0.716  & 0.866   \\
\cline{1-5}
       \multirow{2}{*}{\#recovered}&position RMSE & -1.492  & 1.096  &  0.173  \\
    &jerk RMSE   &  0.306  & 0.537  &    0.570     \\
        \hline \hline
            \end{tabular}
\end{table}

\subsubsection{Grasp-and-Hold Task Metrics}
The following metrics were used for the grasp-and-hold task. 
\paragraph{\#Broken} The number of trials where the object broke. 
\paragraph{\#Dropped} The number of trials where the object slipped and dropped. 
\paragraph{\#Recovered} The number of trials where the object was successfully recovered after slipping.
\paragraph{\#Successful} The number of trials where the object was successfully held without breaking or slipping. 

\subsubsection{Survey Metrics}
Questions one and two asked participants to rate how physically and mentally demanding the task was. The third question asked participants to rate how hurried or rushed the pace of the task was. The fourth question asked participants to rate how insecure, discouraged, irritated, stressed, and annoyed there were. Question five asked participants to rate how successful they were in accomplishing the task. Question six asked participants to rate how hard they had to work to accomplish their level of performance. Question seven asked participants to rate how useful the feedback (Vibration and Squeeze) was, and the last question asked participants to rate how intuitive the control of the virtual gripper was.  

\subsection{Data Analysis}
The data from two participants was excluded from data analysis due to issues that arose during experimentation. The first participant was removed because the breaking thresholds in the grasp-and-hold task were accidentally set up incorrectly. The second participant's data was overwritten due to an error with the data acquisition and control software. The following statistical analyses were performed on the remaining 18 participants (six for each of the three feedback condition order permutations).

All statistical analyses were performed using RStudio(v2.1). Logistic mixed-effects models were used to analyze the trial-by-trial (i.e., binary) broken, dropped, recovered, and successful metrics in the grasp-and-hold task. For these models, participants were treated as random effects, and feedback condition, pulling force, position RMSE, jerk RMSE, and attempt number (i.e., trials 1-3 for each pulling force)  were treated as fixed effects. Total trial number (i.e., 1-24) was treated as a covariate. Total trial number, attempt number, position RMSE, jerk RMSE, and pulling force were analyzed as continuous variables, while haptic feedback condition was analyzed as a categorical variable. A Bonferroni correction was applied for multiple comparisons. 

A linear mixed effects model was used to analyze survey rating results. Condition was the fixed effect and participants were treated as a random effect. 



\section{Results}

\begin{figure}[t!]
    \centering
    \includegraphics[width=0.9\columnwidth]{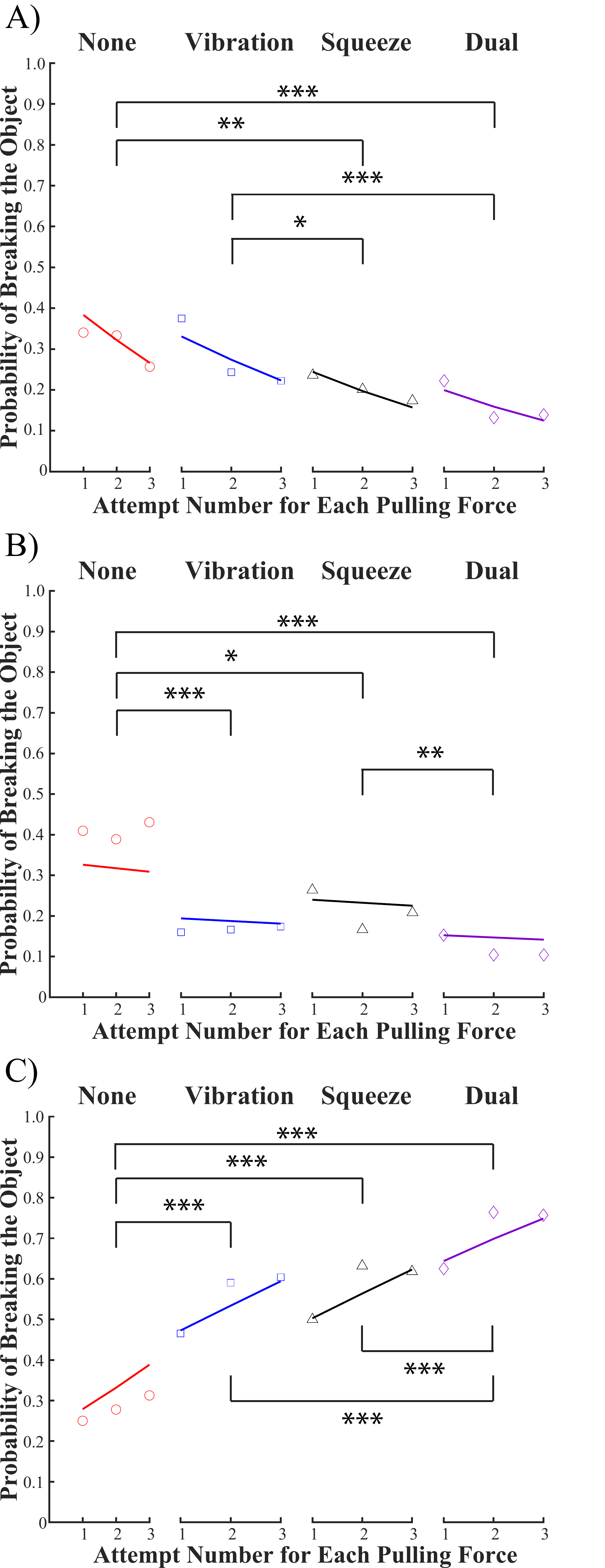}
    \caption{The probability of: (a) breaking, (b) dropping and (c) recovering the object for each trial of pulling force, where the individual data points represent the average for each trial (for all participants in each group), and the solid lines indicate the model’s prediction. * indicates p $<$ 0.05, ** indicates p $<$ 0.01, and *** indicates p $<$ 0.001. }
        \label{fig:probability}
\end{figure}

\subsection{EMG Practice Task}
We first analyzed the effect of participants' performance in the EMG practice task on performance in the grasp-and-hold task (see Table \ref{table:EMG experience}). Results of the models that included $P_{RMSE}$ and $J_{RMSE}$ as fixed effects, found that they were not significant predictors of the broken percentage, dropped percentage, and recovered percentage in the overall data set. To improve modeling accuracy, the statistical models were refined to remove these predictors as fixed effects. The models discussed below still retain feedback condition, pulling force, trial number of each pulling force, and trial number as fixed effects.

\subsection{Grasp-and-Hold Task}

\begin{table}[!t]
\centering
    \small
    \caption{Fixed effect results of the vibration feedback condition, squeezing feedback condition, and dual feedback condition compared to no feedback for \#broken, \#dropped, and \#recovered metrics}
    \label{table:whole data set}
    \renewcommand{\arraystretch}{1.35}
    \begin{tabular}{l l c c c}
        \hline \hline
       metrics & & $\beta$ & \textit{SE} & p \\ [0.5mm]
      \hline 
     \multirow{4}{*}{\#broken}&(Intercept)& 2.286  & 0.419  & \textbf{$<$0.001} \\
    &Vib   & -0.258  & 0.229  &  0.260      \\
    &squeeze &-0.734 &  0.239 & \textbf{$<$0.01}   \\
    &dual  &  -1.021 &  0.238  & \textbf{$<$0.001} \\
    \cline{1-5}
        \multirow{4}{*}{\#dropped}&(Intercept)&-0.031 & 0.506 & 0.952\\
    &Vib   &   -0.851 & 0.249  & \textbf{$<$0.001}\\
    &squeeze &  -0.526  & 0.238  & \textbf{$<$0.05}  \\
    &dual  &  -1.192  & 0.272  & \textbf{$<$0.001}\\
      \cline{1-5}
        \multirow{4}{*}{\#recovered}&(Intercept)& -4.206  & 0.442 & \textbf{$<$0.001}\\
    &Vib   &  1.008 &  0.216 & \textbf{$<$0.001}\\
    &squeeze &  1.154  & 0.217  & \textbf{$<$0.001} \\
    &dual  &  1.855 &  0.223  &  \textbf{$<$0.001}\\
        \hline \hline
            \end{tabular}
\end{table}

When analyzing the results for the grasp-and-hold task, we found that the objected slipped on every single trial for all participants, regardless of feedback condition. This is likely due to the uncertainty caused by the unknown (and invisible) pulling force. Therefore, in our analysis, we omitted the metric \#Successful and instead consider whether object was recovered (i.e., \#Recovered), broke (i.e., \#Broke), or completely slipped (i.e., \#Dropped). Table \ref{table:whole data set} displays the fixed effects result for the three haptic feedback conditions compared to no feedback. 


\subsubsection{\# Broken}
We found a significant fixed effect of pulling force, such that participants were significantly less likely to break the object ($\beta$=-0.234, \textit{SE}=0.031, p$<$0.001) for trials with higher pulling force and correspondingly higher breaking thresholds. There was also a significant fixed effect of attempt number, where increasing attempts with the same object (and same pulling force) decreased the likelihood of object breaks ($\beta$=-0.303, \textit{SE}=0.075, p$<$0.001).
Participants using squeezing feedback ($\beta$=-0.734, \textit{SE}=0.239, p=0.002) and dual feedback ($\beta$=-1.021, \textit{SE}=0.238, p$<$0.001) were significantly less likely to break the object than with no feedback. Compared with vibration feedback, participants using squeezing feedback ($\beta$=-0.476, \textit{SE}=0.171, p=0.015) and dual feedback ($\beta$=-0.763, \textit{SE}=0.179, p$<$0.001) were significantly less likely to break the object. 
There was no significant difference between squeezing feedback and dual feedback. 
See Fig. \ref{fig:probability} for a visualization of these results.

\subsubsection{\# Dropped}
There was a significant fixed effect of the total trial number, with increasing trials decreasing the likelihood of object drops ($\beta$=-0.015, \textit{SE}=0.004, p$<$0.001). 
Participants using vibration feedback ($\beta$=-0.851, \textit{SE}=0.249, p$<$0.001), squeezing feedback ($\beta$=-0.526, \textit{SE}=0.238, p=0.027), and dual feedback ($\beta$=-1.192, \textit{SE}=0.273, p$<$0.001) were significantly less likely to drop the object than with no feedback. Participants using dual feedback ($\beta$=-0.666, \textit{SE}=0.205, p=0.003) were significantly less likely to drop the object than with squeezing feedback. 
There was no significant difference between dual feedback and vibration feedback. 
See Fig. \ref{fig:probability} for a visualization of these results.

\subsubsection{\# Recovered}
We found a significant fixed effect of pulling force, such that participants were significantly more likely to recover the object ($\beta$=0.212, \textit{SE}=0.029, p$<$0.001) for trials with higher pulling force and correspondingly higher breaking thresholds. There was also a significant fixed effect of the total trial number, with increasing trials increasing the likelihood of objects recoveries ($\beta$=0.008, \textit{SE}=0.003, p=0.006). There was also a significant fixed effect of attempt number, where increasing attempts with the same object (i.e, same pulling force and breaking threshold) increasing the likelihood of object recoveries ($\beta$=0.296, \textit{SE}=0.069, p$<$0.001).
Participants using vibration feedback ($\beta$=1.008, \textit{SE}=0.216, p$<$0.001), squeezing feedback ($\beta$=1.154, \textit{SE}=0.217, p$<$0.001) and dual feedback ($\beta$=1.855, \textit{SE}=0.223, p$<$0.001) were significantly more likely to recover the object from slipping than with no feedback. 
Participants using dual feedback were significantly more likely to recover the object from slipping than with vibration feedback ($\beta$=0.847, \textit{SE}=0.159, p$<$0.001) and squeezing feedback ($\beta$=0.700, \textit{SE}=0.159, p$<$0.001). 
See Fig. \ref{fig:probability} for a visualization of these results.


\begin{table*}[htbp]
\caption{Summary of model statistics for survey result}
\makebox[\textwidth][c]{
\begin{tabular}{lccc|ccc|ccc|ccc}
\toprule
& \multicolumn{3}{c}{Intercept (None)} &  \multicolumn{3}{c}{Vibration} & \multicolumn{3}{c}{Squeeze} & \multicolumn{3}{c}{Dual} \\
& $\beta$ & \textit{SE} & p & $\beta$ & \textit{SE} & p & $\beta$ & \textit{SE} & p & $\beta$ & \textit{SE} & p \\
\midrule
Physical effort  
& 4.278 & 0.538 & \textbf{$<$0.001}  
& -0.278 & 0.261 & 0.292 
& -0.556 & 0.261 & \textbf{$<$0.05}    
&-0.444 & 0.261 &  0.094  \\ 
Mental effort  
& 5.778 & 0.573 & \textbf{$<$0.001}  
& -0.889 & 0.369 & \textbf{$<$0.05}  
& -1.111 & 0.369 & \textbf{$<$0.01}    
&-1.333 & 0.369 & \textbf{$<$0.001}   \\
Hurried  
& 3.444 & 0.478 & \textbf{$<$0.001}  
& -0.556 & 0.245 & \textbf{$<$0.05}  
& -0.500 & 0.245 & \textbf{$<$0.05}   
&-0.278 & 0.245 & 0.261  \\
Frustration 
& 3.000 & 0.483 & \textbf{$<$0.001}  
& -0.167 & 0.339 & 0.625  
& -0.389 & 0.339 & 0.256     
&-0.278 & 0.339 & 0.416   \\
Perceived Performance
& 4.833 & 0.485 & \textbf{$<$0.001}  
& 1.278 & 0.454 & \textbf{$<$0.01}  
& 1.333 & 0.454 & \textbf{$<$0.01} 
& 2.278 & 0.454 & \textbf{$<$0.001}   \\
Perceived hardness
& 6.000 & 0.506 & \textbf{$<$0.001}  
& 0.111 & 0.568 & 0.846   
& -0.611 & 0.568 & 0.287   
& -0.222 & 0.568 & 0.697    \\
Intuitive Control
& 6.333 & 0.394 & \textbf{$<$0.001}  
& 0.611 & 0.477 & 0.206 
& 0.989 & 0.485 & \textbf{$<$0.05}   
& 2.333 & 0.477 & \textbf{$<$0.001}   \\

\bottomrule
\end{tabular}
}
\label{tab:survey}
\end{table*}

\subsection{Survey Results} 
Due to the wording of question seven regarding the usefulness of the vibration and squeeze feedback, participants did not consistently answer this question in the single-modality conditions. Table \ref{tab:survey} therefore compares the three haptic feedback conditions with the no feedback condition for the remaining seven post-condition survey questions. Participants in the no feedback condition provided ratings for all survey questions that were significantly different from 0. Participants using squeezing feedback ($\beta$=-0.556, \textit{SE}=0.261, p$<$0.05) rated the task as significantly less physically demanding than with no feedback.
Participants using vibration feedback ($\beta$=-0.889, \textit{SE}=0.369, p$<$0.05), squeezing feedback ($\beta$=-1.111, \textit{SE}=0.369, p$<$0.01 ), and dual feedback ($\beta$=-1.333, \textit{SE}=0.369, p$<$0.001) rated the task as significantly less mentally demanding than with no feedback. 
Participants using vibration feedback ($\beta$=-0.556, \textit{SE}=0.245, p$<$0.05) and squeeze feedback ($\beta$=-0.500, \textit{SE}=0.245, p$<$0.05) rated the pace of the task as less hurried than with no feedback. Participants using vibration feedback ($\beta$=1.278, \textit{SE}=0.454, p$<$0.01), squeezing feedback ($\beta$=1.333, \textit{SE}=0.454, p$<$0.01), and dual feedback ($\beta$=2.278, \textit{SE}=0.454, p$<$0.001) rated their perceived performance as significantly better than with no feedback. Participants rated control of the virtual gripper as significantly more intuitive with squeezing feedback ($\beta$=0.989, \textit{SE}=0.485, p$<$0.01) and dual feedback ($\beta$=2.333, \textit{SE}=0.477, p$<$0.001) than with no feedback. Participants using dual feedback also rated the control of the virtual gripper as significantly more intuitive than with vibration feedback ($\beta$=1.722, \textit{SE}=0.477, p$<$0.001) and squeezing feedback ($\beta$=1.344, \textit{SE}=0.485, p$<$0.01) in a post-hoc test with a Bonferroni correction.

\section{Discussion}
This study investigated whether providing continuous dual-modality haptic feedback can improve myoelectric prosthesis users' ability to perform a dexterous manipulation tasks in which vision was limited. The task, a modified version of the canonical grasp-and-lift task, required participants to grasp and hold a virtual brittle object with a virtual prosthetic gripper. During each trial, the object experienced an increasing load force (termed pulling force) that increased the likelihood of incipient slip. We compared how well participants could perform the task in four separate conditions: no feedback, vibration feedback of incipient slip, squeezing feedback of grip force, and dual (vibration + squeezing) feedback of incipient slip and grip force. Overall, our results suggest that receiving any feedback is better than receiving none, however, dual-modality feedback is far superior than either single-modality feedback in terms of preventing the object from breaking or dropping, even after it started slipping. 

Given the design of our dexterous task, there were two scenarios that lead to task failure. In the first, the object slipped out of the participant's grasp and completely dropped. In the second, the participant squeezed the object too tight and broke it. Both of these failure scenarios were affected by the pulling force that increased throughout each trial. Since the breaking threshold for the object scaled proportional to the final pulling force in each trial (see Eq. \ref{eqn:breaking_threshold}), higher pulling forces resulted in higher breaking thresholds. This relationship, which increased the safe grasping force region, appeared to allow participants to perform the task better regardless of feedback condition, as larger pulling forces lowered the likelihood of object breaks and increased the likelihood of object recoveries after slip. 

Our study design also supported trial-by-trial learning as participants were allowed six grasp attempts with each of the four pulling forces over the course of the 24 trials. Indeed, we observed that the more attempts made in the task overall (i.e., total trial number), the less likely participants were to drop the virtual object and, subsequently, the more likely they were to recover the object once it slipped. In addition, since participants were allowed to attempt each pulling force and breaking threshold three consecutive times before they changed, participants were increasingly likely to recover the object and less likely to break the object with each consecutive grasp-and-hold attempt for a given pulling force and breaking threshold. Both of these results hold true regardless of feedback condition and are in line with previous literature investigating performance in grasp-and-lift and other object manipulation tasks \cite{Flanagan2006ControlTasks,Johansson1988CoordinatedGrip,Johansson1988ProgrammedGrip}. 

In light of the general task performance, the story gets more interesting when considering the impact of the haptic feedback modalities. Our dual-modality feedback provided participants with the necessary slip and grip force information to successfully complete the task. Indeed, both pieces of information proved extremely useful as they decreased the likelihood of breaking or dropping the virtual object, and increased the likelihood of recovering the virtual object compared with no feedback. In addition, the dual-modality feedback led to improved task performance over each of the single-modality feedback approaches in a minimum of two out of the three task metrics. The superiority here of dual-modality feedback is supported by findings in other domains that demonstrated two streams of haptic information are better than one \cite{Walker2015TactileManipulation,Lim2015RoleSurgery,Sullivan2020Multi-SensoryCues,Abiri2019Multi-ModalSurgery,Santos-Carreras2011MultimodalRobotics,Machaca2022TowardsAssessments,Abiri2019ArtificialFeedback}.

That our findings contradict those of Kim et al. \cite{Kim2012HapticAmputees} and Jimenez et al. \cite{Jimenez2014EvaluationLimbs}, likely has to do with our specific focus on tight closed-lop control. As such, our experimental task, metrics, and overall study protocol were designed to tease apart the individual and combined contributions of two different feedback modalities on user's feedforward and feedback control strategies. Also, given Kim et al.'s finding that participants were potentially confused by the simultaneous pressure and shear feedback displayed to the same body part, we conscientiously placed our two feedback modalities in a non-colocated manner. Thus, we can argue that placing two haptic devices on different body parts helped participants better discriminate the different streams of haptic information.

While each single-modality feedback approach allowed for performance better than the no feedback condition, they did not lead to differences in task performance between each other, with the exception of squeezing feedback of grip force, which lowered the likelihood of object breaks compared to vibration feedback of incipient slip. This latter result, however, makes sense give that grip force information, and not slip information, was needed to prevent object breaks. Slip information alone was insufficient to reduce the likelihood of object drops or increase the likelihood of object recoveries compared to grip force information. This observation highlights just how important grip force knowledge is in object manipulation tasks \cite{Kim2012HapticAmputees,Flanagan2006ControlTasks,Johansson1988CoordinatedGrip,Johansson1988ProgrammedGrip}, and why its is the most common form of sensory information investigated in upper-extremity prosthesis research \cite{Thomas2021Sensorimotor-inspiredVision,Thomas2021NeurophysiologicalProstheses,Thomas2019ComparisonProsthesis,Brown2017AnProstheses,Brown2015AnFeedback,Brown2013UnderstandingTask,Witteveen2015VibrotactileUsers,Raveh2018EvaluationParadigm,Chatterjee2008QuantifyingForce,Gillespie2010TowardDevices}. While these findings of improved task performance are generally inline with prior research investigating grasp-and-lift performance with haptic feedback \cite{Brown2013UnderstandingTask,Espinoza2020DesignLifting,Chinello2018DesignGuidance,Walker2014TactileTask,Abbott2021KinestheticProstheses}, they also shed light on the necessity of both slip and grip force information in dexterous task execution.

Given our definitive findings highlighting the benefit of dual-modality feedback over single-modality feedback regarding actual task performance, it is interesting to note that our survey responses don't completely align. while participants rated all three feedback conditions as less mentally demanding than no feedback, only the squeezing feedback condition was rated as being less physically demanding. It is possible that the presence of the vibration feedback increased participants' perception of physical demand in the task, because it consistently alerted them to the object's slipping behavior. Although participants felt less hurried or rushed with vibration feedback or squeezing feedback compared to no feedback, this was no longer the case when these feedback modalities were combined. It is possible that having to tend to two separate streams of haptic information in the condensed trial time window increased the pressure for some participants. 

More inline with the task performance results, participants in all three conditions rated their perceived performance as higher than in the no feedback condition, and rated the gripper control as more intuitive with the conditions featuring squeezing feedback. Most notably, participants rated the dual-modality feedback as more intuitive than either single-modality feedback. This is despite the fact that participants found the task stressful and hard in all four conditions. Thus, it is clear that despite participants' opinions of the task, the feedback allowed them to objectively and subjectively succeed.  

It is worth considering that our EMG practice task results had no measurable impact on participants' ability to perform the grasp-and-hold task. On one hand this suggest that participants' EMG control capability were irrelevant for task success. Still one would think that if control were sufficiently poor, task performance would have to suffer. Therefore, it is likely the case that our task required a level of EMG control proficiency that every participant was able to achieve either innately, or with the help of our pre-task training. It should also be reiterated here that participants utilized EMG to control the velocity of a pair of virtual grippers that did not the carry the inherent electromechanical dynamics of a real prosthetic gripper. Thus, the overall EMG control problem was more tractable than in a real-world scenario. 

While the results of this work are insightful regarding the potential utility of dual-modality feedback for dexterous prosthesis manipulation, there are a few limitations that should be addressed in future research. First, this study utilized a virtual environment approach with non-amputee participants. Future studies should validate these findings with a clinical myoelectric prosthesis under the control of an amputee participant. Second, to minimize the number of condition order permutations needed to properly counterbalance the study, all participants started with the no feedback condition. Given the learning that was observed, it is possible that the no feedback performance could have improved with more practice. While prior research has already demonstrated well the benefit of haptic feedback over no feedback \cite{Brown2013UnderstandingTask,Chatterjee2008QuantifyingForce,Wheeler2010InvestigationSystems, Gillespie2010TowardDevices,Sullivan2021HapticTask,Raveh2018AddingDisturbed,Motamedi2016HapticTasks,Raveh2018EvaluationParadigm,Thomas2019ComparisonProsthesis,Cheng2018Closed-LoopStimulation,Talasaz2017TheSystem,Chinello2018DesignGuidance} future studies could investigate different condition ordering to better understand the longitudinal improvement in each condition. Third, it is possible that our virtual object design made the task too easy by increasing the safe force region at high pulling forces. To counteract this, future investigations could utilize objects that don't have a linear scaling between the pulling force and breaking force. Finally, our EMG practice task didn't seem to have a bearing on performance in the real task. Future investigations could add an EMG assessment task at the end of the study to see how EMG performance changed over the course of the study and if that correlated with real task performance. It will also be necessary to understand the cognitive burden associated with processing multiple streams of haptic information, and if it leads to improved neural efficiency \cite{Thomas2021NeurophysiologicalProstheses} compared to the single-modality feedback approaches. 

\section{Conclusion}
In this study, we investigated the utility of single- and dual-modality haptic feedback for a dexterous object manipulation task and found that dual-modality feedback led to improved task performance and greater perception of intuitive control than either single-modality feedback approach. These findings are of importance to researchers designing next generation prosthetic limbs and researchers in other telerobotic and human--computer interaction applications as it suggest a strong objective and subjective benefit of providing users with the necessary and sufficient haptic information needed to support feedforward and feedback sensorimotor control strategies.

 \section{Acknowledgements}
The authors would like to thank: Sergio Machaca and Garrett Ung, for their support during the virtual task environment setup; Neha Thomas and Leah Jager, for their support during the statistical analysis. 
\ifCLASSOPTIONcaptionsoff
  \newpage
\fi

\bibliography{references}
\bibliographystyle{IEEEtran}

\end{document}